\pdfoutput=1
\documentclass{article}

\newif\ifSubmission
\Submissiontrue
\Submissionfalse

\ifSubmission 

\usepackage{ijcai18}
\usepackage{times}
\usepackage[utf8]{inputenc}
\usepackage[small]{caption}

\else 

\usepackage[a4paper]{geometry}
\usepackage{times}
\usepackage{ijcai-cite}

\fi 

\usepackage[english]{babel}
\usepackage{booktabs}
\usepackage{mathtools}
\usepackage{amssymb}
\usepackage{amsthm}
\usepackage{enumitem}
\usepackage{microtype}
\usepackage{multirow}
\usepackage{thmtools}
\usepackage{tikz}
\usepackage{xcolor}
\usepackage{xspace}
\usetikzlibrary{calc}

\newcommand\e{\mathbin{=}}
\newcommand\n{\mathbin{\neq}}

\newcommand\limp{\supset}

\newcommand\entails{\models}
\newcommand\modelss{\mathrel{{\mathrel|\joinrel\approx}}}
\newcommand\entailss{\modelss}
\newcommand\Names{\mathcal{N}}
\newcommand\Funcs{\mathcal{F}}
\newcommand\card[1]{\left|#1\right|}

\newcommand\KB{\mathrm{KB}}
\newcommand\B[1]{\mathbf{B}_{#1}\mspace{1mu}}
\newcommand\UP[2][]{\mathsf{UP}^{#1}\ifx\uchyph#2\uchyph\else(#2)\fi}
\newcommand\VP[2][]{\mathsf{UP}^{#1}\ifx\uchyph#2\uchyph\else(#2)\fi}

\newcommand{\BigO}{\mathcal{O}}
\newcommand\RES[2]{\mathsf{RES}\ifx\uchyph#1#2\uchyph\else[#1,#2]\fi}
\newcommand\RED[2]{\|#2\|_{\ifx\uchyph#1\uchyph\else#1\fi}}
\newcommand\T{\top}

\newcommand\W{\mathsf{W}}
\newcommand\TRUE{\ensuremath{\normalfont\textsc{true}}}
\newcommand\FALSE{\ensuremath{\normalfont\textsc{false}}}
\newcommand\stdname[1]{{\textrm{#1}}}
\newcommand\func[1]{{\textrm{#1}}}
\newcommand\Sally{\stdname{Sally}}
\newcommand\Frank{\stdname{Frank}}
\newcommand\Fred{\stdname{Fred}}

\newcommand\fatherOf{\func{fatherOf}}
\newcommand\rich{\func{rich}}

\newcommand{\sq}[3]{\mbox{$\medmuskip=#1mu\thickmuskip=#2mu\displaystyle#3$}} 
\makeatletter
\newcommand\thmlinebreak{\leavevmode\@beginparpenalty=10000}
\makeatother
\declaretheoremstyle[headpunct=\ ,bodyfont=\itshape]{thm}
\declaretheoremstyle[headpunct=\ ,bodyfont=\normalfont]{defi}
\declaretheoremstyle[headpunct=.\ ,headfont=\normalfont\itshape,qed=\qedsymbol]{pf}
\declaretheorem[style=thm,name=Theorem]{thm}
\declaretheorem[style=thm,name=Proposition,sibling=thm]{prop}
\declaretheorem[style=thm,name=Lemma,sibling=thm]{lem}
\declaretheorem[style=thm,name=Corollary,sibling=thm]{cor}

\declaretheorem[name={Proof},style=pf,numbered=no]{pf}
\declaretheorem[name={Proof sketch},style=pf,numbered=no]{pfsk}
\SetEnumitemKey{sub}{leftmargin=1em,nosep}
\SetEnumitemKey{nolabel}{label={}}
\newcommand{\complexityclass}[1]{\textup{\textsf{\small #1}}\xspace}
\newcommand{\ptime}{\complexityclass{PTIME}}
\newcommand{\pspace}{\complexityclass{PSPACE}}
\newcommand{\np}{\complexityclass{NP}}
\newcommand{\conp}{\complexityclass{co-NP}}

\newcommand{\fpt}{\complexityclass{FPT}}
\newcommand{\wone}{\complexityclass{W[1]}}
\newcommand{\wtwo}{\complexityclass{W[2]}}
\newcommand{\wsat}{\complexityclass{W[SAT]}}
\newcommand{\wpee}{\complexityclass{W[P]}}

\newcommand{\cowp}{\complexityclass{co-W[P]}}
\newcommand{\aone}{\complexityclass{A[1]}}
\newcommand{\atwo}{\complexityclass{A[2]}}

\newcommand{\awsat}{\complexityclass{AW[SAT]}}
\newcommand{\awp}{\complexityclass{AW[P]}}
\newcommand\EXISTS{\texttt{EXISTS}}
\newcommand\FORALL{\texttt{FORALL}}

\usepackage{acronym}
\acrodef{GG}{Generalized Geography}

\usepackage{todonotes}

\newcommand\waste{\text{\normalfont w}}
\newcommand\order{\text{\normalfont o}}
\newcommand\vertex{{}}
\newcommand\select{{}}

\newcommand{\decisionproblem}[3]{%
\begin{trivlist}
\item
\centerline{%
\begin{tabular}{|r@{\;}p{0.75\columnwidth}|}
\hline
\multicolumn{2}{|l|}{#1} \\
\textit{Instance:} & #2 \\
\textit{Problem:} & #3\\
\hline
\end{tabular}
}
\end{trivlist}
}
\newcommand{\paramdecisionproblem}[4]{
\begin{trivlist}
\item
\centerline{%
\begin{tabular}{|r@{\;}p{0.75\columnwidth}|}
\hline
\multicolumn{2}{|l|}{#1} \\
\textit{Instance:} & #2 \\
\textit{Parameter:} & #3 \\
\textit{Problem:} & #4\\
\hline
\end{tabular}
}
\end{trivlist}
}

\newenvironment{steps}{%
\begin{trivlist}%
\setlength\itemsep{\itemsep*8/10}
}{%
\end{trivlist}%
}%
\newcommand\step[1]{\item[\hskip\labelsep\itshape #1\relax\ifhmode\ifnum\spacefactor>\@m \else.\fi\fi]\ignorespaces}

\ifSubmission 
\title{The Complexity of Limited Belief Reasoning\,---\,The Quantifier-Free Case}
\author{
    Yijia Chen \\ Fudan University \\ Shanghai 201203, China \\ yijiachen@fudan.edu.cn
    \And
    Abdallah Saffidine \\ Australian National University \\ Canberra ACT 0200, Australia \\ abdallah.saffidine@anu.edu.au
    \And
    Christoph Schwering \\ University of New South Wales \\ Sydney NSW 2052, Australia \\ c.schwering@unsw.edu.au
}
\else 
\title{The Complexity of Limited Belief Reasoning\,---\\The Quantifier-Free Case}
\author{
    Yijia Chen \\ Fudan University \\ Shanghai 201203, China \\ yijiachen@fudan.edu.cn
    \and
    Abdallah Saffidine \\ Australian National University \\ Canberra ACT 0200, Australia \\ abdallah.saffidine@anu.edu.au
    \and
    Christoph Schwering \\ University of New South Wales \\ Sydney NSW 2052, Australia \\ c.schwering@unsw.edu.au
}
\fi 

\begin{document}

\maketitle

\begin{abstract}
The classical view of epistemic logic is that an agent knows all the logical consequences of their knowledge base.
This assumption of \emph{logical omniscience} is often unrealistic and makes reasoning computationally intractable.
One approach to avoid logical omniscience is to limit reasoning to a certain \emph{belief level}, which intuitively measures the reasoning ``depth.''

This paper investigates the computational complexity of reasoning with belief levels.
First we show that while reasoning remains tractable if the level is constant, the complexity jumps to \pspace-complete -- that is, beyond classical reasoning -- when the belief level is part of the input.
Then we further refine the picture using parameterized complexity theory to investigate how the belief level and the number of non-logical symbols affect the complexity.
\end{abstract}

\section{Introduction}

The standard way of modeling knowledge and belief\footnote{We use the terms knowledge and belief interchangeably.} in epistemic logic is in terms of \emph{possible worlds}: an agent knows a proposition if and only if it is true in all worlds the agent considers possible.
A side-effect of this model is that agents are assumed to be \emph{logically omniscient}, that is, they know all the consequences of what they know \cite{Hintikka:Omniscience}.

Unfortunately, the assumption of logical omniscience is inappropriate for most resource-bounded agents like humans or robots: it drives up the computational cost of reasoning and is usually far beyond the their capabilities.
Theories of \emph{limited belief} therefore aim to lift the omniscience assumption. 

A number of theories of limited belief have been proposed, predominantly in the 1980s and 1990s \cite{Konolige:DeductiveBelief,Kaplan:ComputationalBelief,Vardi:Omniscience,Fagin:Limited,Levesque:Implicit,PatelSchneider:Implicit,Lakemeyer:Implicit,Delgrande:Implicit}.
A common problem with these approaches is, however, that either their model of limiting belief is too fine-grained or it misses out on simple inferences.

A novel approach to limited belief developed over a series of papers \cite{LLL:SL,LL:LB,LL:ESL,Klassen:Limited,SL:BOL,LL:LBF,Schwering2017} attempts to address this issue.
The basic idea is to stratify beliefs into \emph{belief levels}, where the first one, level $0$, only comprises the explicit beliefs, that is, what is written down expressly in the knowledge base, and higher belief levels $k+1$ draw additional conclusions based on what is believed at level $k$.
Semantically the logic can be characterized using sets of clauses instead of possible worlds, and through \emph{case splits}, that is, by branching on all the values some term can take and propagating the value.

As an example, consider the following knowledge base:
\begin{align*}
& \fatherOf(\Sally) \e \Frank \lor \fatherOf(\Sally) \e \Fred\\
& \fatherOf(\Sally) \e n \limp \rich(n) \e \T \; \text{ for } n \in \{\Frank, \Fred\} \text.
\end{align*}
Here, $\Sally$, $\Frank$, $\Fred$, $\T$ name distinct individuals ($\T$ is an auxiliary name for modeling propositions), whereas $\fatherOf$ and $\rich$ represent functions in the classical sense.
From this knowledge we can deduce $\rich(\Frank) \e \T \lor \rich(\Fred) \e \T$ at level 1 by splitting on all potential fathers of Sally: if Frank is the father, then Frank is rich; if Fred is the father, then he is rich; every other potential father contradicts the first clause.

Logics of limited belief in general and the belief level mechanism in particular aim to provide means of controlling the reasoning effort in a comprehensible and explainable way, as contrasted with using a classical reasoner and terminating it after a timeout, for example.
The rationale behind the belief level approach is that reasoning at small belief levels should be relatively cheap but still sufficient for the average problem a human or a robot faces during their daily operation.
Experiments confirm this hypothesis for the confined domains of Sudoku and Minesweeper \cite{Schwering2017}.

\begin{table}
\centering
\caption{The classification of Limited Belief Reasoning depending on whether the belief level $k$, number of function terms $\card{\Funcs}$, and the number of standard names $\card{\Names}$ are constant, parameters, or input.} \label{table:complexity}
\begin{tabular}{l@{\;\;\;}l@{\;\;\;}lrr}
\toprule
$\card{\Funcs}$ & $k$ & $\card{\Names}$\\
\midrule
\multirow{4}{*}{Input} & Input & --- & \pspace-c & Theorem~\ref{thm:pspace}\\
\cmidrule{2-5}
& \multirow{3}{*}{Param} & Input & \awp-c & Theorem~\ref{thm:awp} \\
\cmidrule{3-5}
& & Param & \multirow{2}{*}{\wpee-c} & \multirow{2}{*}{Proposition~\ref{prop:wp}} \\
& & Const\\
\midrule
\multirow{4}{*}{Param} & Input & \multirow{2}{*}{Input} & \multirow{2}{*}{\cowp-c} & \multirow{2}{*}{Theorem~\ref{thm:cowp}} \\
                       & Param \\
\cmidrule{2-5}
& \multirow{2}{*}{---} & Param & \multirow{2}{*}{\fpt} & \multirow{2}{*}{Proposition~\ref{prop:fpt}} \\
&                      & Const \\
\midrule
--- & Const & --- & \multirow{2}{*}{\ptime} & \multirow{2}{*}{Corollary~\ref{cor:ptime}} \\
Const & --- & --- & &  \\
\bottomrule
\end{tabular}
\end{table}

\subsubsection*{Contribution}

In this paper, we analyze reasoning with belief levels from the perspective of complexity theory.
More precisely, we study the problem of deciding whether a knowledge base entails a query at a certain belief level.
For a constant belief level, the problem is indeed in \ptime\ and hence known to be tractable; the same holds when the knowledge base and query mention only a constant number of function terms.

However, we shall see that if both the belief level and the number of function terms are part of the problem input, then the complexity jumps to \pspace-complete!
This may come as a surprise given that classical, unlimited reasoning is in \conp.
So (large) belief levels appear to make reasoning harder.
Intuitively, the jump is caused by the belief level limiting a possibly scarce resource, namely the number of case splits, which needs to be utilized in an optimal way.

The gap between \ptime\ and \pspace-completeness calls for a more refined analysis, which we carry out using parameterized complexity theory.
We investigate three dimensions of parameters: (1)~the belief level, (2)~the number of function terms mentioned in the reasoning problem (in the above example, the function terms are $\fatherOf(\Sally)$, $\rich(\Frank)$, $\rich(\Fred)$), and (3)~the number of mentioned so-called standard names (in the example, these names are $\Sally$, $\Frank$, $\Fred$, $\T$).
Parameterized complexity theory offers the \complexityclass{W}- and \complexityclass{A}-hierarchies to classify problems between \ptime\ and \np\ and between \np\ and \pspace, respectively.
We locate the parameterized variants of our problem within these hierarchies.

A comprehensive overview of the paper's findings is given in Table~\ref{table:complexity}.
Figure~\ref{fig:zoo} illustrates the relationships between the complexity classes we deal with in this paper.

The paper is structured as follows.
The next section introduces the logic of limited belief and defines the reasoning problem that we shall study.
Section~\ref{sec:ordering} introduces a gadget that we use in several reductions.
Section~\ref{sec:cc} begins the complexity analysis from the perspective of classical complexity theory with \ptime\ and \pspace\ results.
Section~\ref{sec:pc} refines the picture using parameterized complexity theory.
Then we conclude.

\ifSubmission
\else 
This paper is an extended version of \cite{LimitedComplexityIJCAI}.
\fi 

\section{The Logic of Limited Belief} \label{sec:logic}

\begin{figure}
\centering
\tikzstyle{cclass} = [outer sep=0pt,inner sep=0.5ex]
\tikzstyle{included} = [->]
\tikzstyle{corresponds} = [->, densely dashed]
\begin{tikzpicture}[>=latex]
\node[cclass, name=fpt] {\fpt};
\node[cclass, name=wone, above=of fpt, yshift=-4ex] {\wone};
\node[cclass, name=aone, above right=of fpt, yshift=-4ex] {\aone};
\node[cclass, name=wtwo, above=of wone, yshift=-4ex] {\wtwo};
\node[cclass, name=atwo, above right=of wone, yshift=-4ex] {\atwo};
\node[cclass, name=wsat, above=of wtwo, yshift=-0ex] {\wsat};
\node[cclass, name=awsat, above=of atwo, yshift=-0ex] {\awsat};
\node[cclass, name=wp, above=of wsat, yshift=-4ex] {\wpee};
\node[cclass, name=awp, above=of awsat, yshift=-4ex] {\awp};
\node[cclass] at ($(wone)!0.5!(aone)$) {=};
\node[cclass, anchor=center, name=ptime, left=of fpt.center, xshift=-3em] {\rlap{\ptime}\phantom{\ptime}};
\node[cclass, anchor=center, name=pspace, left=of wp.center, xshift=-3em] {\rlap{\pspace}\phantom{\ptime}};
\node[cclass, name=np] at ($(ptime)!0.5!(pspace)$) {\np};
\draw (ptime) edge[included] (np);
\draw (np) edge[included] (pspace);
\draw (fpt) edge[included] (wone);
\draw (wone) edge[included] (wtwo);
\draw (aone) edge[included] (atwo);
\draw (wtwo) edge[included] node[pos=0.45,fill=white,inner sep=1ex]{\ldots} (wsat);
\draw (atwo) edge[included] node[pos=0.45,fill=white,inner sep=1ex]{\ldots} (awsat);
\draw (wsat) edge[included] (wp);
\draw (awsat) edge[included] (awp);
\draw (wtwo) edge[included] (atwo);
\draw (wsat) edge[included] (awsat);
\draw (wp) edge[included] (awp);
\draw (ptime) edge[corresponds] (fpt);
\draw (np) edge[corresponds,bend left=20] ($(wsat.west)$);
\draw (np) edge[corresponds,bend left=20] ($(wp.west)$);
\draw (pspace) edge[corresponds,bend left=35] ($(awsat.north west)!0.5!(awsat.north)$);
\draw (pspace) edge[corresponds,bend left=25] ($(awp.north west)!0.5!(awp.north)$);
\end{tikzpicture}
\caption[Overview of the complexity classes.]{%
    Overview of the classical and parameterized complexity classes relevant for this paper.
    The classes from the \complexityclass{W}-hierarchy include parameterized versions of different natural \np-complete problems; the \complexityclass{A}-hierarchy can be seen as parameterized version of the polynomial hierarchy.
    \smash{%
    \begin{tikzpicture}[>=latex, anchor=base, baseline, every node/.style={inner sep=0,outer sep=0}]
    \node (c1) at (0,0) {$C_1$~~};
    \node (c2) at (3em,0) {~$C_2$};
    \draw (c1) edge[included] (c2);
    \end{tikzpicture}
    }
    means $C_1 \subseteq C_2$, and
    \smash{%
    \begin{tikzpicture}[>=latex, anchor=base, baseline, every node/.style={inner sep=0,outer sep=0}]
    \node (c1) at (0,0) {$C_1$~~};
    \node (c2) at (3em,0) {~$C_2$};
    \draw (c1) edge[corresponds] (c2);
    \end{tikzpicture}
    }
    means that $C_2$ can be seen as a parameterized analogue of $C_1$.
}
\label{fig:zoo}
\end{figure}

In its most recent form, the logic of limited belief is a first-order logic with functions, equality, and epistemic modal operators \cite{LL:LBF,Schwering2017}.
In this paper, we limit our consideration to the \emph{quantifier-free} case.

This section first introduces the syntax and semantics of this logic, and then defines the reasoning problem whose complexity we will study in the remainder of the paper: \emph{if we know $\KB$ explicitly, do we believe $\alpha$ at level $k$?}
The definitions and results of this section are adopted from \cite{Schwering2017} with some minor simplifications to ease the technical treatment; these simplifications do not affect the expressivity or complexity of the reasoning task at hand.

\subsection{The Language}

A term is either a \emph{standard name} (or \emph{name} for short) or a \emph{function term} $f(n_1,\ldots,n_j)$, where $f$ is a function symbol and every $n_i$ is a standard name.
Standard names can be understood as special constants that satisfy the unique-names assumption and an infinitary version of domain closure.
We assume an infinite supply of standard names as well as of function symbols.

A \emph{literal} is an expression of the form $t \e n$ or $\neg t \e n$, where $t$ is a function term and $n$ is a standard name.
A \emph{formula} is a literal or an expression of the form $\neg \alpha$, $(\alpha \lor \beta)$, or $\B k \alpha$, where $\alpha, \beta$ are formulas and $k \geq 0$ is a natural number.
We read $\B k \alpha$ as ``$\alpha$ is known at belief level $k$;'' in case $k = 0$ we also say ``$\alpha$ is known explicitly.''
We use the usual abbreviations $t \n n$, $(\alpha \land \beta)$, $(\alpha \limp \beta)$, and may omit brackets to ease readability. 

A formula without $\B k$ is called \emph{objective}.
\citeauthor{Schwering2017} \shortcite{Schwering2017} has shown that there is a linear Turing reduction from the reasoning problem with nested beliefs to the non-nested case.
Hence to simplify the presentation we henceforth assume that $\alpha$ in $\B k \alpha$ is objective.
As usual, a conjunction of disjunctions of literals is said to be in \emph{conjunctive normal form} (CNF).

\subsection{The Semantics}

The semantics of limited belief is based on clause subsumption, unit propagation, and case splits.
A \emph{clause} is a set of literals.
We abuse notation and identify a non-empty clause $\{\ell_1,\ldots,\ell_j\}$ with the formula $(\ell_1 \lor \ldots \lor \ell_j)$.
In the rest of this paragraph, we implicitly assume that $n, n'$ refer to distinct standard names.
A clause $c_1$ \emph{subsumes} another clause $c_2$ iff for every $t \e n \in c_1$, either $t \e n \in c_2$ or $t \n n' \in c_2$, and for every $t \n n \in c_1$, also $t \n n \in c_2$.
We say two literals $\ell_1,\ell_2$ are \emph{complementary} when they are of the form $t \e n$ and $t \n n$ or of the form $t \e n$ and $t \e n'$.
The \emph{unit propagation} of a clause $c$ and a literal $\ell$ is obtained by removing from $c$ all literals complementary to $\ell$.
For a set of clauses $s$, we let $\UP{s}$ be the closure of $s$ under unit propagation and subsumption.

The \emph{truth relation} $\modelss$ is defined between a formula $\alpha$ and a set of clauses $s$.
Intuitively, $s$ acts as a partial model.
At belief level $0$, $\alpha$ is broken down to clause level and then checked for subsumption by $\VP{s}$.
Higher belief levels allow to branch on a function term $t$ and all its values $n$ and add $t \e n$ to $s$, which may then trigger unit propagation in $\UP{s}$ and thus produce new inferences.
The formal definition is as follows:
\begin{enumerate}[label=\arabic*.,ref=\arabic*,leftmargin=1.2em]
\ifSubmission
\else 
\item \label{sem:c} if $c$ is a clause: \qquad\quad\;\;\;\;\;\,
    $s \modelss c$ \;iff\; $c \in \VP{s}$
\fi 
\item \label{sem:or} if $(\alpha \lor \beta)$ is not a clause: \;
    $s \modelss (\alpha \lor \beta)$ \;iff\; $s \modelss \alpha$ or $s \modelss \beta$
\item \label{sem:neg-or} $s \modelss \neg (\alpha \lor \beta)$ \;iff\; $s \modelss \neg \alpha$ and $s \modelss \neg \beta$
\item \label{sem:neg-neg} $s \modelss \neg \neg \alpha$ \;iff\; $s \modelss \alpha$
\item \label{sem:b0} $s \modelss \B 0 \alpha$ \;iff\; $s \modelss \alpha$
\item \label{sem:bk} $s \modelss \B {k+1} \alpha$ \;iff\;
    for some function term $t$, for all names $n$,\\
    \phantom{$s \modelss \B {k+1} \alpha$ \;iff\;} $s \cup \{t \e n\} \modelss \B k \alpha$
\item \label{sem:neg-b} $s \modelss \neg \B k \alpha$ \;iff\; $s \not\modelss \B k \alpha$
\end{enumerate}
In the remainder, we refer to these definitions as Rules \ref{sem:c}--\ref{sem:neg-b}.
As usual, a formula $\alpha$ is \emph{valid}, written $\entailss \alpha$, iff $s \modelss \alpha$ for every set of clauses $s$.

The belief level $k$ in $\B k \alpha$ specifies the number of case splits, which corresponds to the maximum permitted reasoning effort for proving $\alpha$.
Limited belief is monotonic in the belief level:

\begin{lem} \label{lem:bk-monotonic}
$\entailss \B k \alpha \limp \B {k+1} \alpha$.
\end{lem}

Moreover, belief stabilizes at a high-enough belief level in the following sense:

\begin{lem} \label{lem:bk-stable}
Let $\Funcs$ contain all function terms in $s$ and $\alpha$, and let $k \geq \card{\Funcs}$.
Then $s \modelss \B k \alpha$ iff $s \modelss \B {\card{\Funcs}} \alpha$.
\end{lem}

\subsubsection*{Example}
Let us revisit the example from the introduction to illustrate how the semantics works.
Let $s$ contain the clauses
\begin{align*}
& \fatherOf(\Sally) \e \Frank \lor \fatherOf(\Sally) \e \Fred \\
& \fatherOf(\Sally) \n \Frank \lor \rich(\Frank) \e \T\\
& \fatherOf(\Sally) \n \Fred \lor \rich(\Fred) \e \T
\end{align*}
and let $c$ denote the clause $\rich(\Frank) \e \T \lor \rich(\Fred) \e \T$.
Then $s \modelss \B 1 c$ holds by splitting the cases for Sally's father:
\begin{itemize}
\item $\UP{s \cup \{\fatherOf(\Sally) \e \Frank\}} \ni \rich(\Frank) \e \T$ by unit propagation with the second clause.
\item $\UP{s \cup \{\fatherOf(\Sally) \e \Fred\}} \ni \rich(\Fred) \e \T$ by unit propagation with the third clause.
\item $\UP{s \cup \{\fatherOf(\Sally) \e n\}}$ for $n \notin \{\Frank, \Fred\}$ contains the empty clause by unit propagation with the first clause.
\end{itemize}
In each case, we obtain a clause that subsumes $c$, so for every potential father $n$, $c \in \VP{s \cup \{\fatherOf(\Sally) \e n\}}$.

\subsubsection*{Classical Semantics}
For future reference, we briefly give the classical, ``unlimited'' semantics of objective formulas.
A \emph{world} $w$ is a function from function terms to standard names.
Truth of an objective formula $\alpha$ in a world $w$, written $w \models \alpha$, is defined as follows:
\begin{itemize}
\item $w \models t \e n$ \;iff\; $w(t) = n$
\item $w \models \neg \alpha$ \;iff\; $w \not\models \alpha$
\item $w \models (\alpha \lor \beta)$ \;iff\; $w \models \alpha$ or $w \models \beta$
\end{itemize}
We write $s \entails \alpha$ to say that for all $w$, if $w \models c$ for all $c \in s$, then $w \models \alpha$.
Moreover, we write $\entails \alpha$ for $\emptyset \entails \alpha$.

Limited belief is sound as well as eventually complete with respect to classical semantics in the following sense:

\begin{prop} \label{prop:bk-sound-eventual-complete}
For all finite $s$ and all $\alpha$, there is a (large-enough) belief level $k \geq 0$ such that $s \modelss \B k \alpha$ iff $s \entails \alpha$.
\end{prop}

\begin{pfsk}
Soundness holds because Rule~\ref{sem:bk} branches over all names.
Eventual completeness holds because $k$ can be chosen large enough to split all terms in $s, \alpha$.
\end{pfsk}

\subsection{The Limited Belief Reasoning Problem} \label{sec:logic:lbr}

The fundamental problem of reasoning about limited belief is to decide whether for a given knowledge base $\KB$ and a query $\alpha$, \emph{if $\KB$ is known explicitly, then $\alpha$ is known at belief level $k$}.
In limited belief, $\KB$ is typically assumed to be CNF \cite{LL:LBF}.
The formal definition is:

\decisionproblem{
  Limited Belief Reasoning}{
  Objective formulas $\KB$ and $\alpha$ over function terms $\Funcs$ and standard names $\Names$, $\KB$ in CNF, belief level $k \geq 0$.}{
  Decide whether $\entailss \B 0 \KB \limp \B k \alpha$.}

We shall investigate this problem using classical complexity theory first and then refine the picture using parameterized complexity theory for parameters $k$, $\card{\Funcs}$, $\card{\Names}$.
An overview of the results is in Table~\ref{table:complexity}.

Since the knowledge base in Limited Belief Reasoning is assumed to be in CNF, it corresponds to a unique (modulo $\VP{}$) set of clauses and the problem can be equivalently expressed as a model checking problem:

\begin{lem} \label{lem:lbr-mc}
Let $\KB$ be in CNF with clauses $s = \{c_1, \ldots, c_j\}$.\\
Then $\entailss \B 0 \KB \limp \B k \alpha$ iff $s \modelss \B k \alpha$.
\end{lem}

Thus and by Lemma~\ref{lem:bk-stable} and Proposition~\ref{prop:bk-sound-eventual-complete}, Limited Belief Reasoning is sound and eventually complete with respect to classical reasoning: $\entailss \B 0 \KB \limp \B {\card{\Funcs}} \alpha$ iff $\entails \KB \limp \alpha$.

Finally, the following lemma says that in Rule~\ref{sem:bk} a finite number of function terms and standard names is sufficient.

\begin{lem} \label{lem:bk-finite}
Let $\Funcs$ (resp.\ $\Names$) contain all function terms (resp.\ standard names) in $s, \alpha$, and let $\hat{n} \notin \Names$ be an additional name.
Then $s \modelss \B {k+1} \alpha$ iff for some $t \in \Funcs$, for all $n \in \Names \cup \{\hat{n}\}$, $s \cup \{t \e n\} \modelss \B k \alpha$.
\end{lem}

Together, Lemmas \ref{lem:lbr-mc} and \ref{lem:bk-finite} give rise to a \emph{decision procedure} for Limited Belief Reasoning, which works as follows.
First, the problem is turned into the equivalent model checking problem using Lemma~\ref{lem:lbr-mc}.
Then the procedure applies Lemma~\ref{lem:bk-finite} to reduce the belief level, and finally follows Rules \ref{sem:c}--\ref{sem:b0} to break $\alpha$ down to clause level and check the clauses for subsumption.
It is already known that this procedure runs in time $\BigO(2^k(|\KB|+|\alpha|)^{k+3})$ \cite{Schwering2017}.

\section{Ordering Gadget} \label{sec:ordering}

It is easy to see that the ordering in which terms are split can be relevant.
For example, let $s$ contain the following four clauses:
\begin{align*}
& f \e n \lor g_1 \e n \lor h \e n \quad
& f \n n \lor g_2 \e n \lor h \e n \\
& f \e n \lor g_1 \n n \lor h \e n \quad
& f \n n \lor g_2 \n n \lor h \e n
\end{align*}
Then $s \modelss \B 2 h \e n$ can be proved by splitting $f$ first and then, depending on the value of $f$, splitting $g_1$ or $g_2$ next, but not the other way around.

In this section we construct a gadget that generalizes this idea in order to enforce that a goal formula can only be proved by splitting terms in a certain order (at polynomial cost in space).
This gadget is used repeatedly in the proofs of Sections \ref{sec:cc} and \ref{sec:pc}.
For example, in Theorem~\ref{thm:pspace} we use it to preserve the quantifier ordering of the quantified Boolean formula.

To begin with, the following lemma shows how to make sure that one of the terms from a set $F$ is split no later than at belief level $k$.
We use the notation $[k]$ for $\{1,\ldots,k\}$.

\begin{lem} \label{lem:split-early}
Let $F$ be a non-empty finite set of function terms, and $L$ be a set of literals where every term from $F$ occurs exactly once.
Let $\B k \alpha$ be a formula with $k \geq 1$.
Let $s$ be a set of clauses such that for all $t \notin F$, for some name $n$, $s \cup \{t \e n\} \not\entails \bigvee_{\ell \in L} \ell$ and $s \cup \{t \e n\} \not\entails \bigvee_{\ell \in L} \neg \ell$.

Let \smash{$\ell^\waste_i,\ell^\order_j$} for $i \in [k-1], j \in [k]$ be literals with distinct function terms \smash{$f^\waste_i,f^\order_j$} that do not occur in $s$ or $\alpha$.
Let $c^\order_k$ stand for $\ell^\order_1 \lor \ldots \lor \ell^\order_k$.
Let $s_k$ be the least set that includes $s$ and for every $\ell \in L$ contains the clauses
\begin{itemize}
\item $\neg \ell \lor c^\order_k$
\item $\ell \lor \ell^\waste_1 \lor \ldots \lor \ell^\waste_{k-1} \lor c^\order_k$
\item $\ell \lor \neg \ell^\waste_1 \lor c^\order_k, \;\; \ldots, \;\; \ell \lor \neg \ell^\waste_{k-1} \lor c^\order_k$.
\end{itemize}

Then
\begin{align*}
& \textstyle s_k \modelss \B k (c^\order_k \land (\bigvee_{\ell \in L} \neg \ell \lor \alpha))
\text{\quad iff}\\
& \textstyle \text{for some $t \in F$, for all names $n$, }\\
& \textstyle \qquad\qquad s \cup \{t \e n\} \modelss \B {k-1} (\bigvee_{\ell \in L} \neg \ell \lor \alpha) \text.
\end{align*}
\end{lem}

\begin{pf}
The proof proceeds in four steps.

\begin{steps}
\step{Claim 1}
$s_k \setminus s \not\modelss \B {k-1} c^\order_k$.

\ifSubmission
\else 

\step{Proof of Claim 1}
Let $s'_k = s_k \setminus s$.
The proof is by induction on $k$.
For $k = 1$, it is immediate that $c^\order_k \notin \VP{s'_k}$ and hence $s'_k \not\modelss \B {k-1} c^\order_k$.
For the induction step, suppose $s'_k \not\modelss \B {k-1} c^\order_k$; we show $s'_{k+1} \not\modelss \B k c^\order_{k+1}$.
By Lemma~\ref{lem:bk-finite}, it suffices to show that for all $t \in F \cup \{f^\waste_1,\ldots,f^\waste_k,f^\order_1,\ldots,f^\order_{k+1}\}$, for some $n$, $s'_{k+1} \cup \{t \e n\} \not\modelss \B {k-1} c^\order_{k+1}$.
\begin{itemize}
\item Consider $t \in \{f^\waste_1,\ldots,f^\waste_k\}$.
Without loss of generality, let $t$ is $f^\waste_k$.
Consider a name $n$ such that $t \e n$ is complementary to $\ell^\waste_k$.
Observe that $\VP{s'_k \cup \{t \e n\}} \supseteq \VP{s'_{k+1} \cup \{t \e n\}}$ (*).
By induction, $s'_k \not\modelss \B {k-1} c^\order_k$.
Since $t$ does not occur in $s'_k$ or $c^\order_k$, $s'_k \cup \{t \e n\} \not\modelss \B {k-1} c^\order_{k+1}$.
By (*), $s'_{k+1} \cup \{t \e n\} \not\modelss \B {k-1} c^\order_{k+1}$.

\item Consider $t \in \{f^\order_1,\ldots,f^\order_{k+1}\}$.
Without loss of generality, let $t$ is $f^\order_{k+1}$.
Consider a name $n$ such that $t \e n$ is complementary to $\ell^\order_{k+1}$.
Observe that $\VP{s'_k \cup \{t \e n\}} \supseteq \VP{s'_{k+1} \cup \{t \e n\}}$ (*).
By induction, $s'_k \not\modelss \B {k-1} c^\order_k$.
Since $t$ does not occur in $s'_k$ and does not subsume any literal in $c^\order_{k+1}$, $s'_k \cup \{t \e n\} \not\modelss \B {k-1} c^\order_{k+1}$.
By (*), $s'_{k+1} \cup \{t \e n\} \not\modelss \B {k-1} c^\order_{k+1}$.

\item Consider $t \in F$, and let $\ell \in L$ be the literal whose left-hand side is $t$.
If $\ell$ is an inequality, let $n$ be its right-hand side.
Otherwise, let $n$ be name be such that $t \e n \notin L$.
Then $t \e n$ is not complementary to $\neg \ell$, and since all other literals in $L$ have left-hand sides distinct from $t$, we have that for every $\ell' \in L$, $t \e n$ is not complementary to $\neg \ell'$.
Then $\VP{s'_{k+1} \cup \{t \e n\}} = \VP{s'_{k+1}} \cup \VP{\{t \e n, \ell^\waste_1 \lor \ldots \lor \ell^\waste_{k+1} \lor c^\order_{k+1}, \neg \ell^\waste_1 \lor c^\order_{k+1}, \ldots, \neg \ell^\waste_{k+1} \lor c^\order_{k+1}\}}$.
Using this equivalence, another induction on $k$ shows that $s'_{k+1} \cup \{t \e n\} \not\modelss \B k c^\order_{k+1}$ using the same arguments as in the above two cases for $f^\waste_i, f^\order_j$.
\end{itemize}

\fi 

\step{Claim 2}
For all $t \notin F \cup \{f^\waste_1,\ldots,f^\waste_{k-1},f^\order_1,\ldots,f^\order_k\}$, for some $n$, $s_k \cup \{t \e n\} \not\modelss \B {k-1} c^\order_k$.

\ifSubmission
\else 

\step{Proof of Claim 2}
Let $t \notin F \cup \{f^\waste_1,\ldots,f^\waste_{k-1},f^\order_1,\ldots,f^\order_k\}$.
By assumption, there is some $n$ such that $s \cup \{t \e n\} \not\entails \bigvee_{\ell \in L} (\neg) \ell$ for all $\ell \in L$.
Hence and since $f^\waste_i, f^\order_j$ do not occur in $s$, $s_k \cup \{t \e n\} \modelss \B {k-1} c^\order_k$ iff $s_k \setminus s \modelss \B {k-1} c^\order_k$.
By Claim~1, the claim follows.

\fi 

\step{Claim 3}
For all $t \in F$, for all $n$, $s_k \cup \{t \e n\} \modelss \B {k-1} c^\order_k$.

\step{Proof of Claim 3} Let $t \in F$ and let $n$ be an arbitrary name.
Then for all names $n_1,\ldots,n_k$, $c^\order_k \in \VP{s_k \cup \{t \e n,f^\waste_1 \e n_1,\ldots,f^\waste_{k-1} \e n_{k-1}\}}$.
Thus $s_k \cup \{t \e n\} \modelss \B {k-1} c^\order_k$.

\ifSubmission
\else 

\step{Proof of the lemma}
For the only-if direction suppose that $s_k \modelss \B k (c^\order_k \land (\bigvee_{\ell \in L} \neg \ell \lor \alpha))$.
Then for some $t$, for all $n$, $s_k \cup \{t \e n\} \modelss \B {k-1} (c^\order_k \land (\bigvee_{\ell \in L} \neg \ell \lor \alpha))$.
By Claim~2, $t \in F \cup \{f^\waste_1,\ldots,f^\waste_{k-1},f^\order_1,\ldots,f^\order_k\}$.
Without loss of generality we can assume by Claim~3 and since $f^\waste_i, f^\order_j$ do not occur in $s$ or $\alpha$ that $t \in F$.
Thus for all $n$, $s \cup \{t \e n\} \modelss \B {k-1} (\bigvee_{\ell \in L} \neg \ell \lor \alpha)$.

For the converse direction, let $t \in F$, and suppose that for all $n$, $s \cup \{t \e n\} \modelss \B {k-1} (\bigvee_{\ell \in L} \neg \ell \lor \alpha)$.
Let $n$ be arbitrary.
First suppose that $t \e n$ is complementary to $\neg \ell$ for some $\ell \in L$.
Then $c^\order_k \in \VP{s_k \cup \{t \e n\}}$.
Thus and by assumption, $s_k \cup \{t \e n\} \modelss \B {k-1} (c^\order_k \land (\bigvee_{\ell \in L} \neg \ell \lor \alpha))$.
Now suppose that $t \e n$ is not complementary to any $\neg \ell$ with $\ell \in L$.
There is some $\ell \in L$ whose left-hand side is $t$.
If $\ell$ is of the form $t \e n'$, then $t \e n$ is not complementary to $t \n n'$, so $n, n'$ are distinct and hence $t \e n$ subsumes $t \n n'$.
Otherwise, if $\ell$ is of the form $t \n n'$, then $t \e n$ is not complementary to $t \e n'$, so $n, n'$ are identical and hence $t \e n$ subsumes $t \e n'$.
Hence $\neg \ell \in \VP{s_k \cup \{t \e n\}}$.
Thus and by Claim~3, $s_k \cup \{t \e n\} \modelss \B {k-1} (c^\order_k \land (\bigvee_{\ell \in L} \neg \ell \lor \alpha))$.
Therefore $s_k \modelss \B k (c^\order_k \land (\bigvee_{\ell \in L} \neg \ell \lor \alpha))$.
\qedhere

\fi 

\end{steps}
\end{pf}

The next lemma represents our gadget.
Despite its somewhat intimidating interface, it simply plugs together repeated applications of the previous lemma to completely determine the ordering of splitting terms from $F_1,\ldots,F_l$:

\begin{lem} \label{lem:ordering}
Let $F_1,\ldots,F_l$ be non-empty finite sets of function terms, $F = F_1 \cup \ldots \cup F_l$, and $L_1,\ldots,L_l$ be sets of literals such that every term from $F_i$ occurs exactly once in $L_i$.
Let $\B k \alpha$ be a formula with $k \leq l$.
Let $s$ be a set of clauses such that for all $k \in [l]$, for all $t_l \in F_l,\ldots,t_{k+1} \in F_{k+1},t_k \notin F$, for all $n_l,\ldots,n_{k+1},n_k$, $s \cup s' \not\entails \bigvee_{\ell \in L_k} \ell$ where $s' = \{{t_k \e n_k,}\ldots,t_l \e n_l\}$.

For every set of clauses $s'$, let $s'_i$ and $c^\order_i$ be as in Lemma~\ref{lem:split-early} with respect to $F_i, L_i, \alpha$.
Let $\alpha_0$ be $\alpha$ and $\alpha_i$ for $i > 0$ be $c^\order_i \land (\bigvee_{\ell \in L_i} \neg \ell \lor \alpha_{i-1})$.

Then 
\begin{align*}
& \textstyle ((s_1) \ldots)_k \modelss \B k \alpha_k \text{\quad iff}\\
& \textstyle \text{for some $t_k \in F_k$, for all names $n_k$, \ldots,}\\
& \textstyle \text{for some $t_1 \in F_1$, for all names $n_1$,}\\
& \textstyle \quad\quad s \cup \{t_1 \e n_1,\ldots,t_k \e n_k\} \modelss \bigvee_{\ell \in L_i, i \in [k]} \neg \ell \lor \alpha \text.
\end{align*}
\end{lem}

\ifSubmission
\else 

\begin{pf}
Observe that if $s' \not\entails \bigvee_{\ell \in L_i} (\neg) \ell$, then also $((s'_1) \ldots)_j \not\entails \bigvee_{\ell \in L_i} (\neg) \ell$ (*).
We show by induction on $k$ that for all $t_l \in F_l, \ldots, t_{k+1}$ and for all $n_l,\ldots,n_{k+1}$, $(((s \cup \{t_{k+1} \e n_{k+1},\ldots,t_l \e n_l\})_1) \ldots)_k \modelss \B k \alpha_k$ iff for some $t \in F_k$, for all $n_k$, \ldots, for some $t \in F_1$, for all $n_1$, $s \cup \{t_1 \e n_1,\ldots,t_l \e n_l\} \modelss (\bigvee_{\ell \in L_i, i \in [k]} \neg \ell \lor \alpha)$.
The base case $k = 0$ follows immediately from Rule~\ref{sem:b0}.

For the induction step, consider $k > 0$ and arbitrary $t_l \in F_l,\ldots,t_{k+1} \in F_{k+1}$ and $n_l,\ldots,n_{k+1}$, and suppose the statement holds for $k-1$.
Then
    $((s \cup \{t_{k+1} \e n_{k+1},\ldots,t_l \e n_l\})_1 \ldots)_k \modelss \B k \alpha_k$
iff
    $((s \cup \{t_{k+1} \e n_{k+1},\ldots,t_l \e n_l\})_1 \ldots)_k \modelss \B k (c^\order_{k+1} \land (\bigvee_{\ell \in L_{k+1}} \neg \ell \lor \alpha_k))$
iff (by Lemma~\ref{lem:split-early}, which is applicable by the assumption that for all $t_k \notin F$, for some $n_k$, $s \cup \{t_k \e n_k,\ldots,t_l \e n_l\} \not\entails \bigvee_{\ell \in L_k} (\neg) \ell$ and (*))
    for some $t_k \in F_k$, for all $n_k$, $((s \cup \{t_{k+1} \e n_{k+1},\ldots,t_l \e n_l\})_1 \ldots)_{k-1} \cup \{t_k \e n_k\} \modelss \B {k-1} (\bigvee_{\ell \in L_k} \neg \ell \lor \alpha_{k-1})$
iff
    for some $t_k \in F_k$, for all $n_k$, $((s \cup \{t_k \e n_k,\ldots,t_l \e n_l\})_1 \ldots)_{k-1} \modelss \B {k-1} (\bigvee_{\ell \in L_k} \neg \ell \lor \alpha_{k-1})$
iff (by induction)
    for some $t_k \in F_k$, for all $n_k$, \ldots, for some $t_1 \in F_1$, for all $n_1$, $s \cup \{t_1 \e n_1,\ldots,t_k \e n_k\} \modelss (\bigvee_{\ell \in L_i, i \in [k]} \neg \ell \lor \alpha)$.
\end{pf}

\fi 

\section{Classical Complexity} \label{sec:cc}

This section analyzes the complexity of Limited Belief Reasoning using classical complexity theory.
The next tractability result follows from the decision procedure from Section~\ref{sec:logic:lbr}:

\begin{cor} \label{cor:ptime}
Limited Belief Reasoning with constant $k$ or constant $\card{\Funcs}$ is in \ptime.
\end{cor}

\begin{pf}
The decision procedure runs in time polynomial with degree $k + 3$.
By Lemmas~\ref{lem:bk-monotonic} and \ref{lem:bk-stable}, $k = \card{\Funcs}$ suffices.
\end{pf}

Next, we consider the case where neither $k$ nor $\card{\Funcs}$ is constant.
It comes as no surprise that the complexity then significantly increases with the number of case splits.
Proposition~\ref{prop:bk-sound-eventual-complete} and Lemma~\ref{lem:lbr-mc} showed that Limited Belief Reasoning is sound and eventually complete with respect to classical reasoning.
So clearly, Limited Belief Reasoning must be \conp-hard, and eventual completeness may suggest that it is \conp-complete as well.
However, limiting the number of case splits further adds to the computational complexity: whereas in classical reasoning a decision procedure may ``simply'' split all function terms, a decision procedure for limited belief needs to make sure it makes use of the available case splits in the best possible way.
This leads to the following result:

\begin{thm} \label{thm:pspace}
Limited Belief Reasoning with constant $\card{\Names}$ is \pspace-complete.
The result also holds when $\card{\Names}$ is input.
\end{thm}

\begin{pf}
\begin{steps}
\step{Membership}
The decision procedure from Section~\ref{sec:logic:lbr} runs in space $\BigO(m + k)$ where $m = |\KB| + |\alpha|$, since $\UP{s}$ can be represented in space $\BigO(|s|)$ because minimal clauses suffice.

\step{Hardness}
We reduce from True Quantified Boolean Formula, which is \pspace-complete \cite{Arora:CCAMA}.
The problem input is a fully quantified Boolean formula $Q_k x_k \ldots Q_1 x_1 \psi$ for $Q_i \in \{\forall,\exists\}$ and a propositional formula $\psi$.
Without loss of generality, we assume that $\psi$ mentions negation only in front of variables.
The question is whether this formula evaluates to \TRUE, that is, for all (if $Q_k = \forall$) / some (if $Q_k = \exists$) assignment(s) of $x_k$, \ldots, for all / some assignment(s) of $x_1$, $(x_1,\ldots,x_k)$ satisfies $\psi$.

\ifSubmission
\else 
A truth assignment of $x_1,\ldots,x_j$ is modeled as a vector $M$ of length $k$ with $M_i \in \{\FALSE,\TRUE\}$ to represent the truth assignment of $x_i$.
\fi 
Let $\Names = \{\T, \W\}$ contain two standard names.
Let $\Funcs_\forall = \{f_i \mid Q_i = \forall\}$, $\Funcs_\exists = \{f_i, f'_i \mid Q_i = \exists\}$, $\Funcs = \Funcs_\forall \cup \Funcs_\exists$, where $f_i, f'_i$ are pairwise distinct function terms.
We define a mapping $\ast$ from QBF to limited belief formulas: let $x_i^\ast$ be $f_i \e \T$, let $(\neg x_i)^\ast$ be $f_i \n \T$ if $Q_i = \forall$ and $f'_i \e \T$ if $Q_i = \exists$, let $(\psi_1 \lor \psi_2)^\ast$ be $(\psi_1^\ast \lor \psi_2^\ast)$, and let $(\psi_1 \land \psi_2)^\ast$ be $(\psi_1^\ast \land \psi_2^\ast)$.
For $Q_i = \forall$, let $F_i = \{f_i\}$, $N_i = \{n \mid \text{name $n$ is distinct from $\W$}\}$, and $L_i = \{f_i \n \W\}$; for $Q_i = \exists$, let $F_i = \{f_i,f'_i\}$, $N_i = \{\T\}$, and $L_i = \{f_i = \T, f'_i \e \T\}$.
\ifSubmission
\else 
Let $\alpha$ be $\bigvee_{\ell \in L_i, i \in [k]} \neg \ell \lor \psi^\ast$.
We say a truth assignment $M$ and a set of clauses $s$ are \emph{compatible} iff $s = \{t_1 \e n_1,\ldots,t_k \e n_k\}$ where $t_i \in F_i$, $n_i \in N_i$, and if $Q_i = \forall$, then $M_i = \TRUE$ iff $n_i = \T$, and if $Q_i = \exists$, then $M_i = \TRUE$ iff $t_i = f_i$.
\fi 

The idea is as follows.
Universally quantified $x_i$ are naturally translated to literals $f_i \e \T$ so that the truth values \TRUE\ and \FALSE\ of $x_i$ correspond to $f_i \e \T$ and $f_i \n \T$, respectively.
For existentially quantified $x_i$, positive occurrences of $x_i$ are replaced with $f_i \e \T$ and negative ones with $f'_i \e \T$ so that the truth values \TRUE\ and \FALSE\ of $x_i$ correspond to $f_i \e \T$ and $f'_i \e \T$.
The $f_i$ or $f'_i$ (if $Q_i = \exists$) then need to be split in the appropriate order.

\ifSubmission
\else 

In the remainder of the proof we show that $\phi$ evaluates to \TRUE\ iff for some $t_k \in F_k$, for all $n_i \in N_k$, \ldots, for some $t_1 \in F_1$, for all $n_1 \in N_1$, $\{t_1 \e n_1,\ldots,t_k \e n_k\} \modelss \psi^\ast$ (*).
The right-hand side of (*) holds iff for some $t_k \in F_k$, for all $n_i$, \ldots, for some $t_1 \in F_1$, for all $n_1$, $\{t_1 \e n_1,\ldots,t_k \e n_k\} \modelss \alpha$.
This in turn can be reduced in polynomial time to Limited Belief Reasoning by Lemmas \ref{lem:ordering} and \ref{lem:lbr-mc}, which gives us \pspace-hardness.

We now prove by induction on $j$ that for every truth assignment $M$ of $x_k,\ldots,x_{j+1}$ and every $s$ compatible with $M$, $Q_j x_j \ldots Q_1 x_1 \phi$ evaluates to \TRUE\ under $M$ iff for some $t_j \in F_j$, \ldots, for some $t_1 \in F_1$, for all $n_1 \in N_1$, $s \cup \{t_1 \e n_1,\ldots,t_j \e n_j\} \modelss \psi^\ast$.
For $j = k$ this is identical to (*).
\begin{itemize}
\item For the base case suppose $j = 0$, let $M$ be a truth assignment of $x_k,\ldots,x_1$, and let $s$ be compatible with $M$.
We show that $\psi$ evaluates to \TRUE\ under $M$ iff $s \modelss \psi^\ast$ by subinduction on $\psi$.

For the base case consider a literal.
A positive literal $x_i$ evaluates to \TRUE\ under $M$ iff $M_i = \TRUE$ iff $f_i \e \T \in s$ iff $s \modelss f_i \e \T$ iff $s \modelss x_i^\ast$.
A negative literal $\neg x_i$ with $Q_i = \forall$ evaluates to \TRUE\ under $M$ iff $M_i = \FALSE$ iff $f_i \e n \in s$ for some $n \notin \{\T,\W\}$ iff $f_i \n \T \in \VP{s}$ iff $s \modelss f_i \n \T$ iff $s \modelss (\neg x_i)^\ast$.
A negative literal $\neg x_i$ with $Q_i = \exists$ evaluates to \TRUE\ under $M$ iff $M_i = \FALSE$ iff $f'_i \e \T \in s$ iff $s \modelss f'_i \e \T$ iff $s \modelss (\neg x_i)^\ast$.

The subinduction steps are trivial.

\item For the induction step suppose that the statement holds for $j$, and let $M$ be a truth assignment of $x_k,\ldots,x_j$ and $s$ be compatible with $M$.
We write $M \cdot v$ for the extension of $M$ with value $v \in \{\FALSE,\TRUE\}$ for $x_{j+1}$.

First consider $Q_i = \forall$.
Then $Q_{j+1} x_{j+1} \ldots Q_1 x_1 \psi$ evaluates to \TRUE\ under $M$ iff for every $v \in \{\FALSE,\TRUE\}$, $Q_j x_j \ldots Q_1 x_1 \psi$ evaluates to \TRUE\ under $M \cdot v$ iff (by induction) for every $v \in \{\FALSE,\TRUE\}$, for every $s_v \supseteq s$ that is compatible with $M \cdot v$, for some $t_j \in F_j$, for all $n_j \in N_j$, \ldots, for some $t_1 \in F_1$, for all $n_1 \in N_1$, $s_v \modelss \psi^\ast$ iff for some $t_{j+1} \in F_{j+1}$, for all $n_{j+1} \in N_{j+1}$, \ldots, for some $t_1 \in F_1$, for all $n_1 \in N_1$, $s \modelss \psi^\ast$.

Now consider $Q_i = \exists$.
Then $Q_{j+1} x_{j+1} \ldots Q_1 x_1 \psi$ evaluates to \TRUE\ under $M$ iff for some $v \in \{\FALSE,\TRUE\}$, $Q_j x_j \ldots Q_1 x_1 \psi$ evaluates to \TRUE\ under $M \cdot v$ iff (by induction) for some $v \in \{\FALSE,\TRUE\}$, for the unique $s_v \supseteq s$ that is compatible with $M \cdot v$, for some $t_j \in F_j$, for all $n_j \in N_j$, \ldots, for some $t_1 \in F_1$, for all $n_1 \in N_1$, $s_v \modelss \psi^\ast$ iff for some $t_{j+1} \in F_{j+1}$, for all $n_{j+1} \in N_{j+1}$, \ldots, for some $t_1 \in F_1$, for all $n_1 \in N_1$, $s \modelss \psi^\ast$.
\qedhere
\end{itemize}

\fi 

\end{steps}
\end{pf}

It is noteworthy that this reduction only uses two standard names.
With a more involved reduction, even a single name suffices.
Thus even the propositional case (where an atomic proposition $p$ is simulated by $p \e \T$) is \pspace-complete.

\section{Parameterized Complexity} \label{sec:pc}

The gap between tractability and \pspace-completeness from the previous section calls for a more refined analysis.
In this section we use parameterized complexity theory to investigate how the parameters $k$, $\card{\Funcs}$, and/or $\card{\Names}$ affect the complexity of Limited Belief Reasoning.

While many unparameterized problems can be classified with the classical classes \ptime, \np, or \pspace, parameterized versions of these problems fall into a variety of complexity classes \cite{FlumGrohe}.
The role of \ptime\ in parameterized complexity is taken on by \fpt, which includes problems parameterized by $k$ that are solvable in $f(k) \cdot p(n)$, where $f$ is a computable function and $p$ a polynomial.
Other important parameterized classes come from the \complexityclass{W}- and \complexityclass{A}-hierarchies: the classes $\wone \subseteq \wtwo \subseteq \ldots \subseteq \wsat \subseteq \wpee$ include parameterized versions of different natural \np-complete problems; similarly, the classes $\aone \subseteq \atwo \subseteq \ldots \subseteq \awsat \subseteq \awp$ can be seen as a parameterized version of the polynomial hierarchy.
Figure~\ref{fig:zoo} displays the classes that are relevant for this paper.

Membership in classes of the \complexityclass{W}- and \complexityclass{A}-hierarchies can be shown using machines that restrict the number of nondeterministic steps \cite{ChenFG2005}.
An NRAM is a random access machine with an additional nondeterministic \EXISTS\ instruction, which guesses a natural number less than or equal to a certain register and stores the number in that register.
A problem is in \wpee\ iff it is decidable by an NRAM in $f(k) \cdot p(n)$ steps, at most $g(k)$ of them being nondeterministic, using at most the first $f(k) \cdot p(n)$ registers and only numbers $\leq f(k) \cdot p(n)$.
A problem is in \awp\ iff it is decidable by an ARAM with the same constraints, where an ARAM is an NRAM with an additional nondeterministic \FORALL\ instruction, the dual to \EXISTS.

Hardness in parameterized complexity is shown by way of fpt-reductions, which are reductions computable in time $f(k) \cdot p(n)$ and such that $k' \leq g(k)$, where $k$ and $k'$ are the parameters of the problems reduced from and reduced to, respectively, $n$ is the input size, $f$ and $g$ are computable functions, and $p$ is a polynomial.

Before we start our analysis, we introduce a problem called Quantified Monotone Circuit Satisfiability.
A \emph{circuit} is a directed acyclic graph $(V,E)$ whose vertices are partitioned into input-nodes $X$ of in-degree $0$, not-nodes of in-degree $1$, and- and or-nodes of in-degree $> 0$, and a distinguished output-node $v_0$ of out-degree $0$.
An assignment $S \subseteq X$ sets inputs $S$ to \TRUE\ and the other ones to \FALSE\ and propagates the values to the output node, whose value then determines whether or not $S$ satisfies the circuit.
A \emph{monotone} circuit contains no not-nodes.

\paramdecisionproblem{
  Quantified Monotone Circuit Satisfiability}{
  A monotone circuit $C$ with input-nodes partitioned into sets $X_1,\ldots,X_l$.}{
  $k_1 + \ldots + k_l$}{
  Decide whether for all $S_1 \subseteq X_1$ with $|S_1| = k_1$, for some $S_2 \subseteq X_2$ with $|S_2| = k_2$, \ldots, the assignment $S_1 \cup \ldots \cup S_l$ satisfies $C$.}

The following lemma states \awp-completeness for the problem, which has been claimed elsewhere before without explicit proof \cite{AbrahamsonRF1995}.

\begin{lem} \label{lem:q-m-circuit-sat}
Quantified Monotone Circuit Satisfiability is \awp-complete.
\end{lem}

\begin{pf}
It is sufficient to reduce from Quantified Circuit Satisfiability, which is \awp-complete \cite{FlumGrohe}.
Consider an instance with circuit $C = (V,E)$ and inputs $X_1,\ldots,X_l$.
By De Morgan's laws we can assume the not-nodes are right above the inputs.
\ifSubmission
\else 

Consider a not-node $w \in V$ with incoming edge $\{x,w\} \in E$.
By assumption, $x \in X_i$ for some $i \in [l]$.
Observe that for every assignment $S_i$ of $X_i$, $w$ is \FALSE\ iff $x \notin S_i$ iff $k_i$ variables in $X_i \setminus \{x\}$ are set to \TRUE\ iff $\geq k_i$ variables in $X_i \setminus \{x\}$ are set to \TRUE.
So it suffices to replace $w$ with the following circuit with inputs $X_i$.

Let $X_i \setminus \{x\} = \{x_1,\ldots,x_m\}$.
For every $1 \leq i_1 < i_2 \leq m$ and $t \in [k_i]$, we introduce an or-node $v_{i_1,i_2,t}$ with the meaning that at least $t$ inputs among $x_{i_1},\ldots,x_{i_2}$ are \TRUE, and for every pair of these or-nodes an and-node that takes the pair as inputs.
The inputs of $v_{i_1,i_2,t}$ for $i_2 - i_1 > t$ are the and-nodes of the pairs $v_{i_1,i',t'}$ and $v_{i'+1,i_2,t-t'}$ such that $i_1 \leq i' < i_2$ and $t' \in [t]$, thus representing that $t$ among $x_{i_1},\ldots,x_{i_2}$ are set to \TRUE\ if for some $i'$ and $t'$, $t'$ of $x_{i_1},\ldots,x_{i'}$ and $t'-t$ of $x_{i'+1},\ldots,x_{i_2}$ are \TRUE.
Then we take $v_{1,m,k_i}$ as the output of the desired subcircuit that replaces $w$.

The subcircuit is monotone and can be determined time $\BigO(k_i^2 \cdot m^4)$.
We iterate the procedure to eliminate all negations and thus obtain an fpt-reduction to a monotone circuit.
\qedhere

\fi 

\end{pf}

With this lemma we can establish the complexity of Limited Belief Reasoning parameterized by the belief level:

\begin{thm} \label{thm:awp}
Limited Belief Reasoning with parameter $k$ is \awp-complete.
\end{thm}

\begin{pf}
\begin{steps}
\step{Membership}
We implement the decision procedure from Section~\ref{sec:logic:lbr} using an ARAM.
\ifSubmission
\else 
Model checking at belief level $0$ can clearly be done on a RAM in time $p(m)$ using at most $p(m)$ registers and numbers $\leq p(m)$, where $p$ is a polynomial and $m = |\KB| + |\alpha|$.
At belief level $k > 0$, we select one of the function terms from $\Funcs$ with an \EXISTS\ instruction, and the corresponding name from $\Names \cup \{\hat{n}\}$ using a \FORALL\ instruction.
This amounts to $2 \cdot k$ nondeterministic steps and a total runtime $2 \cdot k \cdot p(m)$, and since $\card{\Funcs} \leq m$ and $\card{\Names} \leq m$, the problem is in \awp.
\fi 

\step{Hardness}
We reduce from Quantified Monotone Circuit Satisfiability, which is \awp-complete by Lemma~\ref{lem:q-m-circuit-sat}.
Let $\sq{1}{2}{C = (V,E)}$ be a monotone circuit with inputs $X_1, \ldots, X_l$.
We say $X_i$ or $x \in X_i$ is \emph{universal} (\emph{existential}) iff $i$ is odd (even).

Let $\Funcs = \{f^\vertex_v \mid v \in V\} \cup \{f^\select_{i,j} \mid X_i \text{ universal}, j \in [k_i]\} \cup \{f^\select_{x,j} \mid x \in X_i \text{ existential}, j \in [k_i]\}$ be function terms.
Let $\Names = \{\T, \W\} \cup \{n^\vertex_x \mid x \in X_i \text{ universal}\}$ be standard names.

\ifSubmission
\else 

The idea is as follows.
A literal $f^\vertex_v \e \T$ represents that node $v$ is \TRUE.
The truth assignment of existential $X_i$ is selected by splitting some $k_i$ of $\{f^\select_{x,j} \mid x \in X_i, j \in [k_i]\}$ and skipping all values but $\T$.
The truth assignment of universal $X_i$ is selected by splitting $f^\select_{i,1},\ldots,f^\select_{i,k_i}$ one after another, whose value denotes which variable should be set to \TRUE: if the value is $n^\vertex_x$, it is $x$.
The truth assignment of existential $X_i$ is selected by splitting one of $\{f^\select_{x,1} \mid x \in X_i\}$, \ldots, one of $\{f^\select_{x,k_i} \mid x \in X_i\}$, whose value is set to $\T$, which then triggers the corresponding $x$ to be set to \TRUE.

For universal $X_i$ and $j \in [k_i]$, let $F_{i,j} = \{f^\select_{i,j}\}$, $N_{i,j} = \{n^\vertex_x \mid x \in X_i\}$, and $L_{i,j} = \{f^\select_{i,j} \n \W\}$.
For universal $X_i$, let $s_i$ be the least set that contains $f^\select_{i,j} \n n^\vertex_x \lor f^\vertex_x \e \T$ for all $j \in [k_i]$ and $x \in X_i$, and $\bigvee_{x \in X_i} f^\select_{i,j} \e n^\vertex_x \lor f^\select_{i,j} \e \W$ for all $j \in [k_i]$.
For existential $X_i$ and $j \in [k_i]$, let $F_{i,j} = \{f^\select_{x,j} \mid x \in X_i\}$, $N_{i,j} = \{\T\}$, and $L_{i,j} = \{f^\select_{x,j} \e \T \mid x \in X_i\}$.
For existential $X_i$, let $s_i$ be the least set that contains $f^\select_{x,j} \n \T \lor f^\vertex_x \e \T$ for all $j \in [k_i]$ and  $x \in X_i$.
Let $s$ be the least set such that $s \supseteq s_i$ for all $i \in [l]$, and that contains $\bigvee_{w \in W} f^\vertex_w \n \T \lor f^\vertex_v \e \T$ for every and-node $v$ and its inputs $W = \{w \mid (w,v) \in E\}$, and $f^\vertex_w \n \T \lor f^\vertex_v \e \T$ for all or-nodes $v$ and all inputs $w$ with $(w,v) \in E$.

In the remainder of the proof we show that for all $S_1 \subseteq X_1$ with $|S_1| \leq k_1$, for some $S_2 \subseteq X_2$ with $|S_2| \leq k_2$, \ldots, the truth assignment $S_1 \cup \ldots \cup S_l$ satisfies $C$ iff for some $t_{1,1} \in F_{1,1}$, for all $n_{1,1} \in N_{1,1}$, \ldots, for some $t_{1,k_1} \in F_{1,k_1}$, for all $n_{1,k_1} \in N_{1,k_1}$, \ldots, for some $t_{l,1} \in F_{l,1}$, for all $n_{l,1} \in N_{l,1}$, \ldots, for some $t_{l,k_l} \in F_{l,k_l}$, for all $n_{l,k_l} \in N_{l,k_l}$, $s \cup \{t_{1,1} \e n_{1,1},\ldots,t_{1,k_1} \e n_{1,k_1},\ldots,t_{l,1} \e n_{l,1},\ldots,t_{l,k_l} \e n_{l,k_l}\} \modelss f^\vertex_{v_0} \e \T$ (*).
The left-hand side of (*) is equivalent to the Quantified Monotone Circuit Satisfiability problem due to the circuit's monotonicity.
The right-hand side of (*) holds iff for some $t_{1,1} \in F_{1,1}$, for all $n_{1,1}$, \ldots, for some $t_{1,k_1} \in F_{1,k_1}$, for all $n_{1,k_1}$, \ldots, for some $t_{l,1} \in F_{l,1}$, for all $n_{l,1}$, \ldots, for some $t_{l,k_l} \in F_{l,k_l}$, for all $n_{l,k_l}$, $s \modelss \bigvee_{\ell \in L_{i,j}, i \in [l], [j] \in [k_i]} \neg \ell \lor f^\vertex_{v_0} \e \T$ since $s$ contains clauses that restrict the domain of $f^\select_{i,j}$ for universal $X_i$ to $\{n^\vertex_x \mid x \in X_i\} \cup \{\W\}$.
This in turn fpt-reduces to Limited Belief Reasoning by Lemma~\ref{lem:ordering}, which is applicable here since the $F_{i,j}$ are mutually disjoint, and Lemma~\ref{lem:lbr-mc}, which gives us \awp-hardness.

We first define when a truth assignment $S_i \subseteq X_i$ and a set of clauses $s'$ are \emph{compatible}:
if $X_i$ is universal, then $S_i, s'$ are compatible iff $|S_i| = |s'|$ and for every $x \in S_i$, for some $j \in [k_i]$, $f^\select_{i,j} \e n^\vertex_x \in s'$;
if $X_i$ is existential, then $S_i, s'$ are compatible iff $|S_i| = |s'|$ and for every $x \in S_i$, for some $j \in [k_i]$, $f^\select_{x,j} \e \T \in s'$.
We now prove by induction on $j \leq l$ that for a given truth assignment $S_1 \cup \ldots \cup S_j$ the following statement: for all (if $j+1$ is odd) / some (otherwise) $S_{j+1} \subseteq X_{j+1}$ with $|S_{j+1}| \leq k_{j+1}$, for some / all $S_{j+2} \subseteq X_{j+2}$ with $|S_{j+2}| \leq k_{j+2}$, \ldots, $S_1 \cup \ldots \cup S_l$ satisfies $C$ iff for all compatible $s'_1,\ldots,s'_j$, for some $t_{j+1,1} \in F_{j+1,1}$, for all $n_{j+1,1} \in N_{j+1,1}$, \ldots, for some $t_{j+1,k_{j+1}} \in F_{j+1,k_{j+1}}$, for all $n_{j+1,k_{j+1}} \in N_{j+1,k_{j+1}}$, \ldots, $t_{l,1} \in F_{l,1}$, for all $n_{l,1} \in N_{l,1}$, \ldots, for some $t_{l,k_l} \in F_{l,k_l}$, for all $n_{l,k_l} \in N_{l,k_l}$, $s \cup s'_1 \cup \ldots \cup s'_j \cup \{t_{1,1} \e n_{1,1},\ldots,t_{1,k_1} \e n_{1,k_1},\ldots,t_{l,1} \e n_{l,1},\ldots,t_{l,k_l} \e n_{l,k_l}\} \modelss f^\vertex_{v_0} \e \T$.
For $j = 0$ this is identical to (*).

\begin{itemize}
\item For the base case $j = l$ we show by subinduction on the depth of $C$ that $v \in V$ is satisfied iff $f^\vertex_x \e \T \in \UP{s \cup s'_1 \cup \ldots \cup s'_j}$.
The base case holds since $S_i$ and $s'_i$ are compatible and by construction of $s$.
The subinduction step follows immediately by construction of $s$.

\item For the induction step suppose the claim holds for $j+1$.

First suppose $j+1$ is odd.
Then
    for all $S_{j+1} \subseteq X_{j+1}$ with $|S_{j+1}| \leq k_{j+1}$, for some $S_{j+2} \subseteq X_{j+2}$ with $|S_{j+2}| \leq k_{j+2}$, \ldots, $S_1 \cup \ldots \cup S_l$ satisfies $C$
iff (by induction)
    for all $S_{j+1} \subseteq X_{j+1}$ with $|S_{j+1}| \leq k_{j+1}$,
    for all compatible $s'_1,\ldots,s'_{j+1}$,
    some $t_{j+2,1} \in F_{j+2,1}$, for all $n_{j+2,1} \in N_{j+2,1}$, \ldots, for some $t_{j+2,k_{j+2}} \in F_{j+2,k_2}$, for all $n_{j+2,k_2} \in N_{j+1,k_{j+2}}$, \ldots, $t_{l,1} \in F_{l,1}$, for all $n_{l,1} \in N_{l,1}$, \ldots, for some $t_{l,k_l} \in F_{l,k_l}$, for all $n_{l,k_l} \in N_{l,k_l}$, $s \cup s'_1 \cup \ldots \cup s'_j \cup \{t_{1,1} \e n_{1,1},\ldots,t_{1,k_1} \e n_{1,k_1},\ldots,t_{l,1} \e n_{l,1},\ldots,t_{l,k_l} \e n_{l,k_l}\} \modelss f^\vertex_{v_0} \e \T$
iff
    for all compatible $s'_1,\ldots,s'_j$,
    for some $t_{j+1,1} \in F_{j+1,1}$, for all $n_{j+1,1} \in N_{j+1,1}$, \ldots, for some $t_{j+2,k_{j+2}} \in F_{j+2,k_2}$, for all $n_{j+2,k_2} \in N_{j+1,k_{j+2}}$, \ldots, $t_{l,1} \in F_{l,1}$, for all $n_{l,1} \in N_{l,1}$, \ldots, for some $t_{l,k_l} \in F_{l,k_l}$, for all $n_{l,k_l} \in N_{l,k_l}$, $s \cup s'_1 \cup \ldots \cup s'_j \cup \{t_{1,1} \e n_{1,1},\ldots,t_{1,k_1} \e n_{1,k_1},\ldots,t_{l,1} \e n_{l,1},\ldots,t_{l,k_l} \e n_{l,k_l}\} \modelss f^\vertex_{v_0} \e \T$.

The case for even $j+1$ is analogous.
\qedhere
\end{itemize}

\fi 

\end{steps}
\end{pf}

Membership in \awp\ is quite natural due to the alternation of existential and universal quantifications of case splits in Lemma~\ref{lem:bk-finite}.
When the number of standard names $\card{\Names}$ becomes a parameter as well, this gives us leverage to replace the nondeterministic \FORALL\ steps that select the standard names with a simple loops.
It is therefore not surprising that Limited Belief Reasoning parameterized by $k$ and $\card{\Names}$ is in \wpee, the hardest \np-analogue of the \complexityclass{W}-hierarchy.
The following result shows that the problem is in fact \wpee-complete:

\begin{prop} \label{prop:wp}
Limited Belief Reasoning with parameters $k$ and $\card{\Names}$ is \wpee-hard.
The result also holds when $\card{\Names}$ is constant.
\end{prop}

\begin{pf}
\begin{steps}
\step{Membership}
We build an NRAM.
For $k = 0$, it behaves like the ARAM in Theorem~\ref{thm:awp}.
For $k > 0$, we select a function term from $\Funcs$ with \EXISTS\ and loop over all names in $\Names \cup \{\hat{n}\}$.
\ifSubmission
\else 
This amounts to $(\card{\Names} + k)^k$ nondeterministic steps and total runtime $(\card{\Names} + k)^k \cdot p(m)$, so the problem is in \wpee.
\fi 

\ifSubmission
\else 

\step{Hardness}
We reduce from Weighted Monotone Circuit Satisfiability, which is \wpee-complete \cite{AbrahamsonRF1995}.
Weighted Circuit Satisfiability corresponds to Quantified Weighted Circuit Satisfiability with only existentially quantified variables, and the reduction is identical to one from Theorem~\ref{thm:awp} for a single block of existential variables.
The reduction uses only a single name $\T$.
\qedhere

\fi 

\end{steps}
\end{pf}

Next we consider the case where $\card{\Funcs}$ becomes a parameter.
The below theorem specifies \cowp-completeness:

\begin{thm} \label{thm:cowp}
Limited Belief Reasoning with parameters $k$ and $\card{\Funcs}$ is \cowp-complete.
The result also holds when $k$ is input.
\end{thm}

\begin{pf}
\begin{steps}
\step{Membership}
We show that the co-problem is in \wpee\ using an NRAM that finds a falsifying assignment of names for all split terms.
As in Theorem~\ref{thm:awp}, the case $k = 0$ is straightforward.
For $k > 0$ we loop over all function terms in $\Funcs$ and for each we select a standard name from $\Names \cup \{\hat{n}\}$ with \EXISTS.
\ifSubmission
\else 
This amounts to $\smash{\card{\Funcs}^k}$ nondeterministic steps, which by Lemma~\ref{lem:bk-stable} can be generalized to $\smash{\card{\Funcs}^{\card{\Funcs}}}$.
The total runtime is hence $\smash{\card{\Funcs}^{\card{\Funcs}}} \cdot p(m)$, so the problem is in \cowp.
\fi 

\step{Hardness}
We reduce from the complement of Weighted Anti-Monotone Circuit Satisfiability, which is \wpee-complete \cite{FlumGrohe}.
A circuit is anti-monotone when all inputs have out-degree $1$ and feed into a not-node and there are no other not-nodes except those on top of some input.
Let $C = (V,E)$ be an anti-monotone circuit with inputs $X$.
For every $x \in X$ we denote the associated not-node by $v_x$.

Let $\Funcs = \{f^\select_i \mid i \in [k]\} \cup \{f\}$ be function terms.
Let $\Names = \{n^\vertex_v \mid v \in V \setminus X\} \cup \{\W\}$ be standard names.

The idea is to represent that a node $v$ is set to $\FALSE$ by $f \n n^\vertex_v$.
The truth assignment is selected by splitting $f^\select_1,\ldots,f^\select_k$.
Truth of an input $x$ is represented by $f^\select_i \e n^\vertex_{v_x}$ for some $i \in [k]$, which triggers $f \n n^\vertex_{n_x}$; these values are propagated to the output node, so that $f \n n^\vertex_{v_0}$ indicates that the circuit is falsified.

\ifSubmission
\else 

For every $i \in [k]$, let $L_i = \{f^\select_i \n \W\}$ and let $s_i$ be the least set that contains $f^\select_i \n n^\vertex_{v_x} \lor f \n n^\vertex_{v_x}$ for every $x \in X$, and $\bigvee_{x \in X} f^\select_i \e n^\vertex_{v_x} \lor f^\select_i \e \W$.
Let $s$ be the least set such that $s \supseteq s_i$ for all $i \in [k]$, and $f^\select_i \n n^\vertex_{v_x} \lor f^\select_j \n n^\vertex_{v_x}$ for every $i, j \in [k]$ with $i \neq j$ and $x \in X$, and $\bigvee_{w \in W} f \e n^\vertex_w \lor f \n n^\vertex_v$ for every or-node $v$ and its inputs $W = \{w \mid (w,v) \in E\}$, and $f \e n^\vertex_w \lor f \n n^\vertex_v$ for all and-nodes $v$ and all inputs $w$ with $(w,v) \in E$.

Let $N = \{n^\vertex_{v_x} \mid x \in X\}$.
In the remainder of the proof we show that every truth assignment $S \subseteq X$ with $|S| = k$ falsifies $C$ iff for all $n_1,\ldots,n_k \in N$, $s \cup \{f^\select_1 \e n_1,\ldots,f^\select_k \e n_k\} \modelss f \n n^\vertex_{v_0}$ (*).
The right-hand side of (*) holds iff for all $n_1,\ldots,n_k$, $s \cup \{f^\select_1 \e n_1,\ldots,f^\select_k \e n_k\} \modelss \bigvee_{\ell \in L_i, i \in [k]} \neg \ell \lor f \n n^\vertex_{v_0}$ since $s$ contains clauses that restrict the domain of $f^\select_i$ to $N \cup \{\W\}$.
This in turn fpt-reduces to Limited Belief Reasoning by Lemmas \ref{lem:ordering} and \ref{lem:lbr-mc}, which gives us \cowp-hardness.

We first define that a truth assignment $S \subseteq X$ and a set of clauses $s'$ are \emph{compatible} iff $|S| = |s'|$ and for every $x \in S$, for some $i \in [k]$, $f^\select_i \e n^\vertex_{v_x} \in s'$.
We now prove (*).

For the only-if direction suppose that every $S \subseteq X$ falsifies $C$ and consider some $n_1,\ldots,n_k \in N$.
Let $s' = \{f^\select_1 \e n_1,\ldots,f^\select_k \e n_k\}$.
If for some $i \neq j$, $n_i = n_j$, then by construction, $s \cup s' \modelss f \n n^\vertex_{v_0}$.
Hence $x \in S$ iff for some $i$, $f^\select_i \e n^\vertex_{v_x} \in s'$.
Thus and by construction of $s$, $S$ falsifies $v_x$ iff $f \neq n^\vertex_{v_x} \in \UP{s \cup s'}$.
By induction on the depth of $C$, $f \n n^\vertex_{v_0}$.

For the if direction suppose that for all $n_1,\ldots,n_k \in N$, $s \cup \{f^\select_1 \e n_1,\ldots,f^\select_k \e n_k\} \modelss f \n n^\vertex_{v_0}$ and consider some $S \subseteq X$.
Choose $n_1,\ldots,n_k$ such that $x \in S$ iff for some $i$, $f^\select_i \e n^\vertex_{v_x} \in s'$.
Then by construction of $s$, $S$ falsifies $v_x$ iff $f \neq n^\vertex_{v_x} \in \UP{s \cup s'}$.
By induction on the depth of $C$, $S$ falsifies $v_x$ iff $f \neq n^\vertex_{v_x} \in \UP{s \cup s'}$.
\qedhere

\fi 

\end{steps}
\end{pf}

Finally, the only remaining case is when Limited Belief Reasoning is parameterized by both $\card{\Funcs}$ and $\card{\Names}$:

\begin{prop} \label{prop:fpt}
Limited Belief Reasoning with parameters $\card{\Funcs}$ and $\card{\Names}$ is in \fpt.
The result also holds when $\card{\Names}$ is constant.
\end{prop}

\begin{pf}
\ifSubmission
\else 
The decision procedure for belief level $k$ corresponds to a tree of height $2 \cdot k$ with alternating branching factors $\card{\Funcs}$ and $\card{\Names} + k$.
The runtime of the decision procedure is hence $(\card{\Funcs} \cdot (\card{\Names} + k))^k \cdot p(m)$ for some polynomial $p$ and $m = |\KB| + |\alpha|$.
By Lemma~\ref{lem:bk-stable}, this generalizes to $(\card{\Funcs} \cdot (\card{\Names} + \card{\Funcs}))^{\card{\Funcs}} \cdot p(m)$.
\fi 
\end{pf}

\section{Conclusion} \label{sec:conclusion}

We have analyzed the complexity of Limited Belief Reasoning.
While tractable for constant belief levels, the complexity jumps to \pspace-complete in the general case.
Using parameterized complexity theory, we showed how parameterized versions of the problem populate the space between these two extremes.

We believe our findings are relevant to the future development of the theory of limited belief.
In particular, the insight that the limited belief level can actually increase the computational cost should be considered in future versions.

In light of \pspace-completeness, one might implement a reasoning system using an off-the-shelf QBF-solver. 
Also, limited belief and the available reasoning system \cite{Schwering2017} may be suitable as a modeling language for other problems in \pspace.

So far, we have only considered Limited Belief Reasoning without first-order quantification; lifting this restriction would be a natural next step.
Moreover, additional parameters could be studied, for example, parameters exploiting the structure of the knowledge base and the query, like backdoors \cite{Gaspers:Backdoors}.

Another interesting question is whether our findings carry over to other approaches to resource-bounded reasoning, such as \cite{Agostino:InformationalView}, which uses a similar splitting technique.


\begin{thebibliography}{}

\bibitem[\protect\citeauthoryear{Abrahamson \bgroup \em et al.\egroup
  }{1995}]{AbrahamsonRF1995}
Karl~A. Abrahamson, Rodney~G. Downey, and Michael~R. Fellows.
\newblock Fixed-parameter tractability and completeness {IV}: On completeness
  for {W[P]} and {PSPACE} analogues.
\newblock {\em Annals of Pure and Applied Logic}, 73(3):235--276, 1995.

\bibitem[\protect\citeauthoryear{Arora and Barak}{2009}]{Arora:CCAMA}
Sanjeev Arora and Boaz Barak.
\newblock {\em Computational complexity: a modern approach}.
\newblock Cambridge University Press, 2009.

\bibitem[\protect\citeauthoryear{Chen \bgroup \em et al.\egroup
  }{2005}]{ChenFG2005}
Yijia Chen, J{\"o}rg Flum, and Martin Grohe.
\newblock Machine-based methods in parameterized complexity theory.
\newblock {\em Theoretical Computer Science}, 339(2-3):167--199, 2005.

\bibitem[\protect\citeauthoryear{Chen \bgroup \em et al.\egroup
  }{2018}]{LimitedComplexityIJCAI}
Yijia Chen, Abdallah Saffidine, and Christoph Schwering.
\newblock The complexity of limited belief reasoning\,---\,the quantifier-free
  case.
\newblock In {\em 27th International Joint Conference on Artificial
  Intelligence (IJCAI)}, 2018.
\newblock To appear.

\bibitem[\protect\citeauthoryear{D'Agostino}{2015}]{Agostino:InformationalView}
Marcello D'Agostino.
\newblock An informational view of classical logic.
\newblock {\em Journal of Theoretical Computer Science}, 606(C), 2015.

\bibitem[\protect\citeauthoryear{Delgrande}{1995}]{Delgrande:Implicit}
James~P. Delgrande.
\newblock A framework for logics of explicit belief.
\newblock {\em Computational Intelligence}, 11(1), 1995.

\bibitem[\protect\citeauthoryear{Fagin and Halpern}{1987}]{Fagin:Limited}
Ronald Fagin and Joseph Halpern.
\newblock Belief, awareness, and limited reasoning.
\newblock {\em Artificial intelligence}, 34(1), 1987.

\bibitem[\protect\citeauthoryear{Flum and Grohe}{2006}]{FlumGrohe}
J{\"o}rg Flum and Martin Grohe.
\newblock {\em Parameterized Complexity Theory}.
\newblock Springer-Verlag, 2006.

\bibitem[\protect\citeauthoryear{Gaspers and Szeider}{2012}]{Gaspers:Backdoors}
Serge Gaspers and Stefan Szeider.
\newblock Backdoors to satisfaction.
\newblock In {\em The Multivariate Algorithmic Revolution and Beyond}, pages
  287--317. Springer, 2012.

\bibitem[\protect\citeauthoryear{Hintikka}{1975}]{Hintikka:Omniscience}
Jaakko Hintikka.
\newblock Impossible possible worlds vindicated.
\newblock {\em Journal of Philosophical Logic}, 4(4), 1975.

\bibitem[\protect\citeauthoryear{Kaplan and
  Schubert}{2000}]{Kaplan:ComputationalBelief}
Aaron~N. Kaplan and Lenhart~K. Schubert.
\newblock A computational model of belief.
\newblock {\em Artificial Intelligence}, 120(1):119--160, 2000.

\bibitem[\protect\citeauthoryear{Klassen \bgroup \em et al.\egroup
  }{2015}]{Klassen:Limited}
Toryn~Q. Klassen, Sheila~A. McIlraith, and Hector~J. Levesque.
\newblock Towards tractable inference for resource-bounded agents.
\newblock In {\em 12th International Symposium on Logical Formalizations of
  Commonsense Reasoning}, 2015.

\bibitem[\protect\citeauthoryear{Konolige}{1986}]{Konolige:DeductiveBelief}
Kurt Konolige.
\newblock A deduction model of belief.
\newblock {\em Research notes in Artificial Intelligence}, 1986.

\bibitem[\protect\citeauthoryear{Lakemeyer and Levesque}{2013}]{LL:LB}
Gerhard Lakemeyer and Hector~J. Levesque.
\newblock Decidable reasoning in a logic of limited belief with introspection
  and unknown individuals.
\newblock In {\em 23rd International Joint Conference on Artificial
  Intelligence (IJCAI)}, 2013.

\bibitem[\protect\citeauthoryear{Lakemeyer and Levesque}{2014}]{LL:ESL}
Gerhard Lakemeyer and Hector~J. Levesque.
\newblock Decidable reasoning in a fragment of the epistemic situation
  calculus.
\newblock In {\em 14th International Conference on Principles of Knowledge
  Representation and Reasoning (KR)}, 2014.

\bibitem[\protect\citeauthoryear{Lakemeyer and Levesque}{2016}]{LL:LBF}
Gerhard Lakemeyer and Hector~J. Levesque.
\newblock Decidable reasoning in a logic of limited belief with function
  symbols.
\newblock In {\em 15th International Conference on Principles of Knowledge
  Representation and Reasoning (KR)}, 2016.

\bibitem[\protect\citeauthoryear{Lakemeyer}{1994}]{Lakemeyer:Implicit}
Gerhard Lakemeyer.
\newblock Limited reasoning in first-order knowledge bases.
\newblock {\em Artificial Intelligence}, 71(2), 1994.

\bibitem[\protect\citeauthoryear{Levesque}{1984}]{Levesque:Implicit}
Hector~J. Levesque.
\newblock A logic of implicit and explicit belief.
\newblock In {\em 4th National Conference on Artificial Intelligence (AAAI)},
  1984.

\bibitem[\protect\citeauthoryear{Liu \bgroup \em et al.\egroup }{2004}]{LLL:SL}
Yongmei Liu, Gerhard Lakemeyer, and Hector~J. Levesque.
\newblock A logic of limited belief for reasoning with disjunctive information.
\newblock In {\em 9th International Conference on Principles of Knowledge
  Representation and Reasoning (KR)}, 2004.

\bibitem[\protect\citeauthoryear{Patel-Schneider}{1990}]{PatelSchneider:Implicit}
Peter~F. Patel-Schneider.
\newblock A decidable first-order logic for knowledge representation.
\newblock {\em Journal of Automated Reasoning}, 6(4), 1990.

\bibitem[\protect\citeauthoryear{Schwering and Lakemeyer}{2016}]{SL:BOL}
Christoph Schwering and Gerhard Lakemeyer.
\newblock Decidable reasoning in a first-order logic of limited conditional
  belief.
\newblock In {\em 22nd European Conference on Artificial Intelligence (ECAI)},
  2016.

\bibitem[\protect\citeauthoryear{Schwering}{2017}]{Schwering2017}
Christoph Schwering.
\newblock A reasoning system for a first-order logic of limited belief.
\newblock In {\em 26th International Joint Conference on Artificial
  Intelligence (IJCAI)}, 2017.

\bibitem[\protect\citeauthoryear{Vardi}{1986}]{Vardi:Omniscience}
Moshe~Y. Vardi.
\newblock On epistemic logic and logical omniscience.
\newblock In {\em 1st Conference on Theoretical Aspects of Reasoning about
  Knowledge (TARK)}, 1986.

\end{thebibliography}

\end{document}